\documentclass[letterpaper]{article}
\let\pdfinfoPrimitive\pdfinfo
\usepackage{aaai2027}
\usepackage[hyphens]{url}
\usepackage{graphicx}
\usepackage{natbib}
\usepackage{caption}
\usepackage{booktabs}
\usepackage{amsmath}
\usepackage{amssymb}
\usepackage{multirow}
\usepackage{tabularx}
\usepackage{array}
\usepackage{makecell}
\usepackage{mathtools}

\newlength{\dynlhswidth}
\newcolumntype{Y}{>{\centering\arraybackslash}X}
\pdftrailerid{}
\pdfinfoPrimitive{
/TemplateVersion (2027.1)
/Creator ()
/Producer ()
}

\title{
\makebox[\textwidth][c]{%
  \raisebox{-0.35\height}{%
    \includegraphics[width=1.15cm]{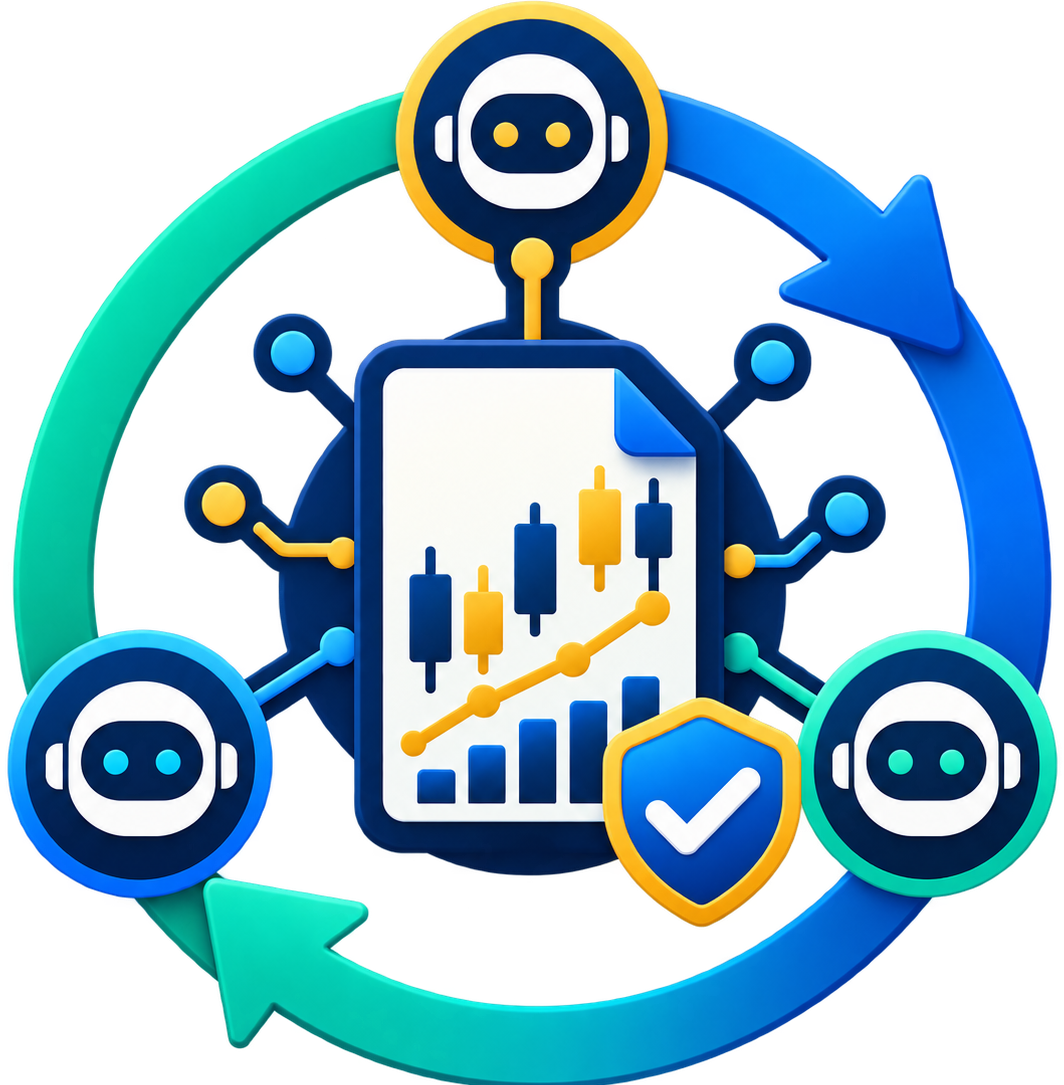}%
  }%
  \hspace{0.2em}
  \parbox{0.82\textwidth}{%
    \centering
    FinCUABuild: Can Agents Build Reliable Benchmarks for Dynamic Financial Computer Use?
  }%
}
}

\author{
Jingpu Yang\textsuperscript{\rm 1,3,4}\thanks{~~Equal contribution.},
Fengxian Ji\textsuperscript{\rm 1,2,3,4}\footnotemark[1],
Jinri Guo\textsuperscript{\rm 3},
Tianhao Li\textsuperscript{\rm 3},
Qian Jiang\textsuperscript{\rm 3},
Fan Zhang\textsuperscript{\rm 2}\\
Min Peng\textsuperscript{\rm 1},
Qianqian Xie\textsuperscript{1}\thanks{Corresponding author.},
Preslav Nakov\textsuperscript{\rm 2},
Zhuohan Xie\textsuperscript{2}\footnotemark[2] \\
}
\affiliations{
\renewcommand{\arraystretch}{1.15}
\begin{tabular}{c}
\textsuperscript{\rm 1}School of Artificial Intelligence, Wuhan University, \textsuperscript{\rm 2}MBZUAI, \textsuperscript{\rm 3}Northeastern University, \textsuperscript{\rm 4}Zhongguancun Academy
\end{tabular}
\\
\texttt{jingpuyang290@gmail.com}, \texttt{202316187@stu.neu.edu.cn}\\
\texttt{\{fengxian.ji, fan.zhang, zhuohan.xie, preslav.nakov\}@mbzuai.ac.ae} \\
\texttt{\{pengm, xieq\}@whu.edu.cn, 202510490@stu.neuq.edu.cn, GreatKowal@outlook.com}
}
\nocopyright

\begin{document}

\maketitle

\begin{abstract}
Financial scenarios are diverse and complex, spanning varying data conditions, tool configurations, and workflows. Yet existing CUA, Computer-Using Agent, evaluation tasks remain largely manually constructed, limiting scalable coverage of real-world financial scenarios.
Then, Can agents autonomously construct diverse CUA evaluation tasks for financial scenarios? Evaluating this capability poses three key challenges: scenario coverage of construction requests, fair comparison across construction methods, and reliable assessment of generated task quality.
To sovle these, we introduce \textit{\textbf{FinCUABuildBench}}, a benchmark for evaluating financial CUA task construction, featuring: (i) 576 construction requests covering 24 financial workflows and three types of runtime variation; (ii) standardized input, budget, and output specifications; and (iii) a task qualification mechanism based on execution tests and quality checks.
We further introduce \textit{\textbf{FinCUABuildAgent}}, a multi-agent system for automatically constructing dynamic financial CUA evaluation tasks. It consists of three modules that jointly construct tasks, environments, and validators.
On FinCUABuildBench, under the same model backbone, existing agent-based construction methods achieve strict qualification rates of only 1.3–8.3\%, while FinCUABuildAgent reaches 31.3\%. Downstream evaluations further show that the constructed tasks can effectively differentiate CUA task-execution capabilities.
These results demonstrate that agents can autonomously construct financial CUA tasks with meaningful evaluation value, offering a practical path toward broader evaluation coverage in financial scenarios.
Code: https://github.com/FengxianJi/FinCUABuild
\end{abstract}

\section{Introduction}

Evaluation of computer-use agents (CUAs) has expanded to real-world web environments, desktop interactions, and enterprise workflows \cite{zhou2024webarenarealisticwebenvironment,drouin2024workarenacapablewebagents,xie2024osworldbenchmarkingmultimodalagents,ji2026style}. Existing interactive benchmarks further provide controllable application states, dynamically instantiated tasks, and policy-constrained tool interactions, laying the foundation for evaluating complex workflows \cite{trivedi2024appworld,rawles2025androidworld,yao2025taubench,zhou2026fincards,elbadry2026sahm}. In finance, workflows involve cross-file retrieval, spreadsheet operations, computation, and reporting, and must cover diverse data conditions, tool combinations, and business processes. SpreadsheetBench and Finch respectively explore real-world spreadsheet manipulation and enterprise-level financial workflow evaluation. Notably, even with LLM-assisted task mining, Finch still required over 700 hours of expert annotation \cite{ma2024spreadsheetbenchchallengingrealworld,dong-etal-2026-finch}. Such construction costs hinder the scalable expansion of evaluation tasks and motivate our central question: Can agents autonomously construct CUA evaluation tasks that cover diverse financial scenarios?

Existing work still has two key limitations in constructing dynamic financial CUA tasks. First, there is no unified benchmark for evaluating task construction capabilities, as existing financial benchmarks mainly assess agents on completing predefined tasks \cite{chen2022finqadatasetnumericalreasoning,zhu2021tatqa,ma2024spreadsheetbenchchallengingrealworld,dong-etal-2026-finch}. Second, there is no construction agent tailored to this setting. Although prior automated construction methods have explored goal-driven generation, multi-agent collaboration, and task synthesis, they provide limited support for jointly constructing financial evidence, application states, dynamic events, and verification rules \cite{li2025autobencher,butt2024benchagents,xie2026agentsynth}. Therefore, we study both how to evaluate financial CUA task construction capabilities and how to improve them.

\begin{figure*}[!t]
    \centering
    \includegraphics[width=1\textwidth]{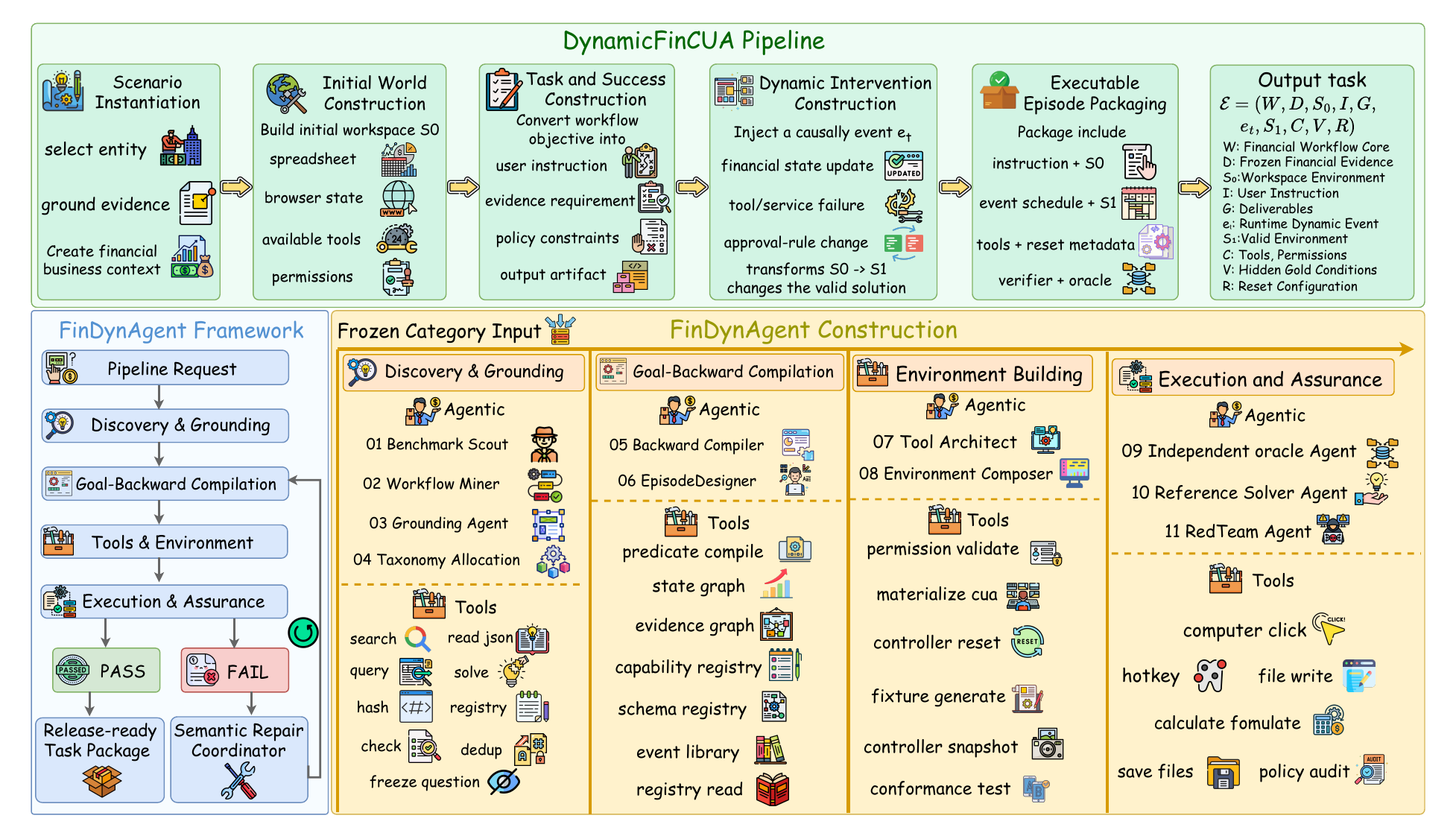}
    \caption{
Overview of FinCUABuildAgent. The upper panel shows the construction of a
FinCUABuildBench episode, while the lower panel groups eleven specialist
agents and their representative tools by stage. Labels 01--11 denote
agents $A_{01}$--$A_{11}$ in the text; shortened agent names are used
for compactness.}
    \label{fig:FinCUABuildAgent-overview}
\end{figure*}

Addressing these two issues requires tackling challenges in both construction-capability evaluation and automated task construction. For evaluation, (i) construction requests should cover diverse financial workflows and runtime variations, rather than increasing diversity through superficial paraphrasing; (ii) differences in input information, tool access, and resource budgets can confound method comparisons, requiring standardized evaluation conditions; and (iii) a plausible task description does not guarantee an executable environment, valid runtime events, or correct verification, so the quality of the complete task package must be assessed. For construction, (i) abstract capability targets must be translated into concrete dependencies among evidence, states, and actions to form executable tasks; (ii) tasks and verifiers may jointly accept incorrect outcomes, requiring independent correctness auditing; and (iii) errors can propagate across dependencies among tasks, environments, and verifiers, requiring failure localization and targeted repair of affected artifacts.

To address these challenges, we introduce \textit{\textbf{FinCUABuildBench}}, shown as Fig.~\ref{fig:FinCUABuildAgent-overview}(a), a benchmark for evaluating the construction of dynamic financial CUA tasks. For scenario coverage, the benchmark includes 576 construction requests derived from 24 financial workflow categories, 192 workflow templates, and three types of runtime variation: state updates, tool failures, and policy changes. For fair comparison, it standardizes construction inputs, resource budgets, and output specifications so that different methods are evaluated under comparable conditions. For quality assessment, it combines financial evidence checking, reference execution, reset-and-replay testing, dynamic-event validation, and verifier testing to determine whether a generated task is qualified. We further introduce \textit{\textbf{FinCUABuildAgen}}, shown as Fig.~\ref{fig:FinCUABuildAgent-overview}(b), which uses capability-dependency compilation to translate target capabilities into task requirements, independent quality auditing to inspect tasks and verifiers, and targeted feedback-driven repair to resolve cross-stage errors, jointly constructing tasks, environments, and verifiers.
Under the same model backbone, FinCUABuildAgent achieves a strict qualification rate of 31.3\%, outperforming general-purpose construction agents at 1.3–8.3\%.
The 144 qualified tasks reveal differences in downstream CUAs' dynamic adaptation capabilities, while the qualification rate on 90 held-out construction requests is 13.3\%.
These results support the evaluation value of automated task construction, while also indicating substantial room for improved generalization.

Our main contributions and findings are:
(1) A benchmark for benchmark construction.
We introduce FinCUABuildBench, a 576-slot construction benchmark spanning 24 workflow categories, 192 WorkflowCores, and three runtime-dynamic profiles. Its unified episode contract evaluates groundedness, replayability, verifier reliability, dynamicity, capability necessity, safety, and redundancy;
(2) A capability-guided multi-agent framework.
We propose FinCUABuildAgent, which models dependencies among states, evidence, operators, and tools. Specialized agents jointly ground financial sources, build environments and runtime events, construct verifiers, and iteratively test and repair auditable CUA episodes;
(3) A controlled empirical evaluation.
Under the same model backbone, FinCUABuildAgent achieves a 31.3\% strict qualification rate, compared with 1.3–8.3\% for general-purpose construction agents. The 144 qualified tasks distinguish downstream CUAs in dynamic adaptation, while a 13.3\% rate on 90 held-out requests indicates cross-workflow transferability.

\section{Related Work}

\paragraph{Financial benchmarks and workflow agents.}
Financial benchmarks first emphasized report-based question answering and numerical reasoning through FinQA, TAT-QA, ConvFinQA, FinanceBench, and MultiHiertt \cite{chen2022finqadatasetnumericalreasoning, zhu2021tatqa, chen2022convfinqaexploringchainnumerical, islam2023financebenchnewbenchmarkfinancial, zhao2022multihiertt}, then expanded to knowledge-intensive reasoning and financial data analysis in FinanceMATH, FinBen, and FinDABench \cite{zhao2024financemath,liu2025findabench,xie2024finben}
. Spreadsheet-oriented work moves toward executable artifact manipulation: SheetCopilot and SheetAgent study spreadsheet control, while SpreadsheetBench and Finch cover realistic workbook and enterprise finance and accounting tasks \cite{li2023sheetcopilot, chen2025sheetagent, ma2024spreadsheetbenchchallengingrealworld, dong-etal-2026-finch}. Recent financial-agent work further targets professional spreadsheets, financial research, banking workflows, MCP-based tool use, and execution-grounded safety \cite{kundurthy2026bluefinbenchmarkingllmagents, wang2026bigfinancebenchworkflowgroundedbenchmarkfinancialresearch, lau2026bankertoolbenchevaluatingaiagents, zhu2026finmcpbench, yang2026finvault}. FinCUABuildBench evaluates the construction process itself: whether a method can produce a replayable episode whose evidence, application state, tools, post-start event, reset logic, and verifier remain jointly consistent.

\paragraph{Computer-use agents and stateful tool use.}
General CUA benchmarks span realistic web and desktop interaction \cite{zhou2024webarenarealisticwebenvironment, xie2024osworldbenchmarkingmultimodalagents}, compositional enterprise work and professional data workflows \cite{drouin2024workarenacapablewebagents, boisvert2024workarenaplus, cao2024spider2v}, and controllable application and mobile environments \cite{trivedi2024appworld,rawles2025androidworld}. ToolSandbox and $\tau$-bench add stateful, multi-turn, or policy-constrained tool interaction \cite{lu2025toolsandbox,yao2025taubench}. These studies evaluate agent execution in supplied task environments. FinCUABuildBench instead targets the synthesis of a complete package comprising the environment, task, event, and verifier.

\paragraph{Automated benchmark construction.}
Automated construction ranges from Dynabench's human--model loop to graph-generated or combinatorial evaluations in DyVal and SKILL-MIX \cite{kiela2021dynabench,zhu2024dyval,yu2024skillmix}. AutoBencher optimizes declared desiderata, BENCHAGENTS coordinates planning, generation, verification, and evaluation, and the Benchmark Agent preprint adds design, grounding, and allocation stages \cite{li2025autobencher,butt2024benchagents,xiong2026benchmarkagent}. APIGen-MT generates verified multi-turn task blueprints, while EnvScaler and AgentSynth synthesize tool environments, terminal validators, or verified long-horizon tasks \cite{prabhakar2025apigenmt,song2026envscaler,xie2026finmmeval_lab,zhang2026finreporting}. We instantiate AutoBencher, BENCHAGENTS, and Benchmark Agent as matched construction baselines. FinCUABuildAgent further couples financial grounding, cross-artifact state, deterministic post-start events, reset and replay semantics, and mutation-tested hidden verifiers within a single release contract.

\section{FinCUABuildBench}

\subsection{Problem Formulation}

We study the automatic construction of benchmarks for dynamic financial computer use. Let $N^\star$ be the number of preregistered target slots, indexed by $i\in\{1,\ldots,N^\star\}$. The construction request for slot $i$ is defined as

\begin{equation}
\mathcal{I}_i
=
(w_i,\kappa_i,\mathcal{S},\mathcal{A},\mathcal{U},\Gamma),
\label{eq:construction-request}
\end{equation}
where $w_i$ specifies the financial workflow and $\kappa_i$ specifies
the target capability. The sets $\mathcal{S}$, $\mathcal{A}$, and
$\mathcal{U}$ provide the admissible sources, application environments,
and builder-side tools, respectively, while $\Gamma$ contains the
safety, privacy, resource, and release constraints.

Let $\mathcal{M}$ denote the set of benchmark-construction methods. Given $\mathcal{I}_i$, a method $m\in\mathcal{M}$ generates a candidate episode
\begin{equation}
x_i
=
m(\mathcal{I}_i)
=
(u_i,D_i,s_{0,i},G_i,T_i,\varepsilon_i,V_i,R_i,\Pi_i).
\label{eq:candidate-episode}
\end{equation}
Here, $u_i$ is the user-visible instruction; $D_i$ is a versioned source
bundle; $s_{0,i}$ is the resettable initial state; $G_i$ contains the
terminal goal predicates; $T_i$ is the evaluated agent's ToolGraph;
$\varepsilon_i$ is a deterministic runtime-event program; $V_i$ is a
hidden verifier; $R_i$ is a reset and replay manifest; and $\Pi_i$
specifies permissions and forbidden effects.

A candidate is qualified only if it satisfies every mandatory
constraint in $\Gamma$ and passes every preregistered episode-level
hard gate: L-Acc, Trace, Exec, verifier pass, Dyn, CapMatch, and
nonduplication. The strict conjunction is formalized in
Eq.~\eqref{eq:qualification-indicator}. H-Acc and CapCov are reported
as complementary sample-level and portfolio-level diagnostics and do
not enter the episode-level release indicator.

\begin{equation}
\begin{aligned}
q_i = \mathbf{1}\!\bigl[&
\Gamma_i \land \mathrm{LAcc}_i \land \mathrm{Trace}_i \\
&\land \mathrm{Exec}_i \land \mathrm{VPass}_i
\land \mathrm{Dyn}_i \\
&\land \mathrm{CapMatch}_i \land \neg\mathrm{Dup}_i
\bigr].
\end{aligned}
\label{eq:qualification-indicator}
\end{equation}

\begin{figure*}[t]
    \centering
    \includegraphics[width=1\textwidth]{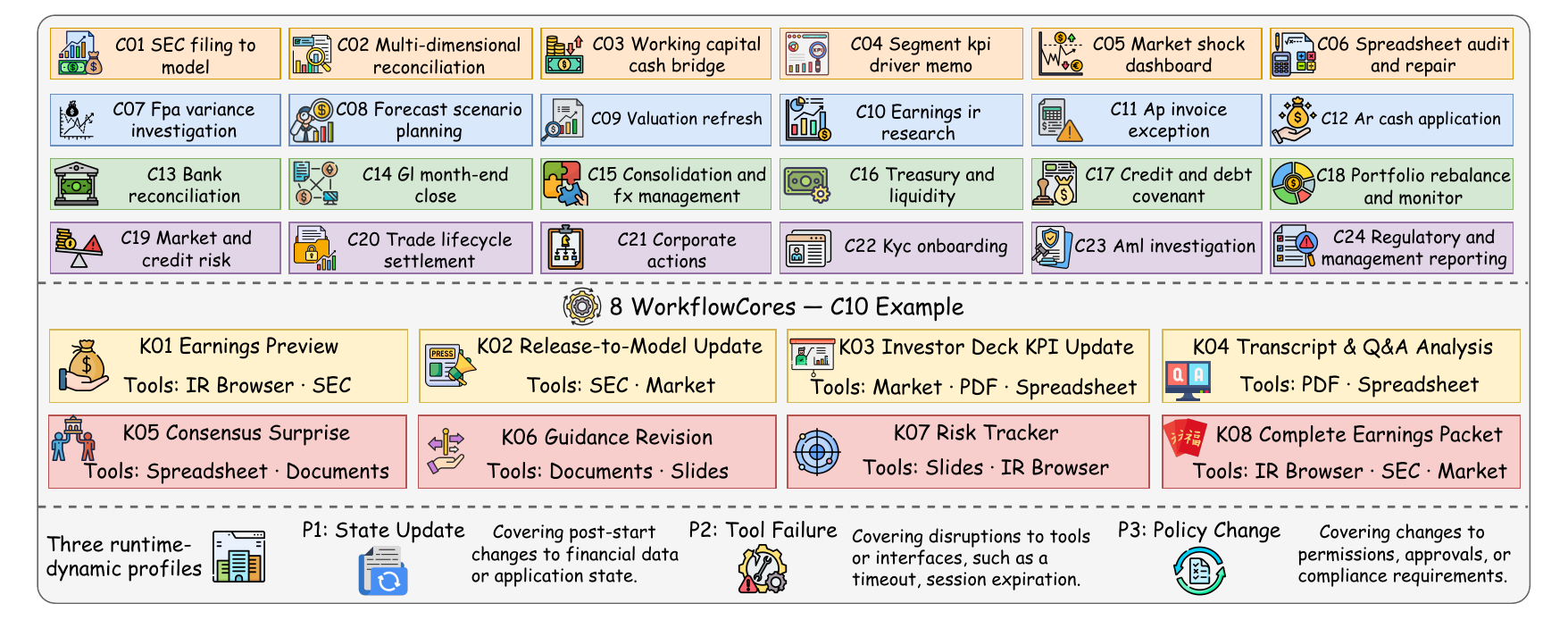}
    \caption{
\textbf{The FinCUABuildBench task space.} 24 financial workflow categories, each with eight semantically distinct WorkflowCores, instantiated under three runtime-dynamic profiles.}
    \label{fig:FinCUABuildBench-task-space}
\end{figure*}

\subsection{FinCUABuildBench Construction}

FinCUABuildBench is designed to evaluate whether a construction method can generate complete dynamic financial computer-use episodes. It covers 24 financial workflow categories spanning reporting and analysis, planning and accounting operations, markets, risk and treasury, and compliance and governance. For each category, we define eight semantically distinct WorkflowCores, each specified by its business objective, source family, artifact contract, and target capability path. Each WorkflowCore is instantiated under the three runtime-dynamic profiles summarized in Figure~\ref{fig:FinCUABuildBench-task-space}, defining 576 preregistered target slots in total.

\begin{itemize}
    \item \textbf{P1: State Update}, covering post-start changes to financial data or application state, such as an amended filing, refreshed market value, changed invoice status, collaborator edit, or updated ledger record;

    \item \textbf{P2: Tool Failure}, covering disruptions to tools or interfaces, such as a timeout, session expiration, delayed recalculation, unavailable primary tool, partial write, or changed interface state;

    \item \textbf{P3: Policy Change}, covering changes to permissions, approvals, or compliance requirements, such as an approval rejection, modified threshold, permission restriction, compliance hold, or newly introduced instruction that conflicts with the applicable policy.
\end{itemize}

Each runtime event is injected after execution begins, is deterministically replayable, changes at least one action required for success, and leaves at least one feasible completion path. To prevent workflow-level leakage, the P1--P3 instances of the same WorkflowCore are assigned to the same split. Each episode follows the contract in Eq.~\eqref{eq:candidate-episode}. Public workflows use versioned, hashed sources, whereas enterprise workflows use privacy-safe synthetic fixtures with cross-file consistency checks. The evaluated agent observes only the task inputs, workspace, and permitted tools; the event controller, reference goal state, oracle, verifier, and scoring logic remain evaluator-private.

Candidate episodes undergo iterative generation, audit, and repair, and only those satisfying the strict release criterion enter the qualified pool. For downstream evaluation, we sample two qualified episodes from each of the $24\times3$ category--profile cells, yielding 144 episodes. All selected episodes pass independent execution, deterministic replay, verifier, event-validity, and safety audits, and the frozen reference solver succeeds on all of them. The subset is fixed before any downstream-agent results are observed.

\subsection{Automation-Point Verification}
\label{sec:automation-point-verification}

FinCUABuildBench evaluates executions through \emph{automation points},
which are programmatically verifiable semantic checkpoints in a financial workflow. For each episode $i$, we organize the task objective, artifact contract, evidence requirements, and runtime event into an automation-point DAG
$\mathcal{G}^{\mathrm{AP}}_i=(\mathcal{V}_i,\mathcal{E}_i)$,
where $\mathcal{V}_i$ contains the automation points and
$\mathcal{E}_i$ encodes their prerequisite and event-order dependencies.
A point may represent grounding a financial fact, updating an artifact,
transferring information across tools, or adapting to a runtime event.
Low-level interface actions are excluded.

For condition
$\chi\in\{\mathrm{static},\mathrm{dynamic}\}$, let
$\mathcal{X}_{\chi}$ be the evaluated episode indices and let
$\mathcal{V}_{i,\chi}\subseteq\mathcal{V}_i$ be the nonempty set of points
applicable to episode $i$. For run $r\in\{1,\ldots,n_i\}$ of agent $a$,
where $n_i\geq1$ is the number of repeated runs, let
$\omega_{i,\chi}^{a,r}$ contain the observed application states, produced
artifacts, and execution and evidence trace. For
$j\in\mathcal{V}_{i,\chi}$, let $\phi_{ij,\chi}$ and
$\operatorname{dep}_{ij,\chi}$ be Boolean predicates that check the semantic
outcome and its dependency constraints, respectively. The binary node-pass
indicator is

\begin{equation}
z_{ij,\chi}^{a,r}
=
\mathbf{1}\!\left[
\phi_{ij,\chi}\!\left(\omega_{i,\chi}^{a,r}\right)
\land
\operatorname{dep}_{ij,\chi}
\!\left(\omega_{i,\chi}^{a,r}\right)
\right],
\label{eq:automation-point-pass}
\end{equation}

where $\mathbf{1}[P]$ equals one when $P$ is true and zero otherwise.
Because verification uses semantic outcomes, different valid tools and
action orders may satisfy the same automation point.

Let $\beta_{ij,\chi}>0$ be preregistered point weights satisfying
$\sum_{j\in\mathcal{V}_{i,\chi}}\beta_{ij,\chi}=1$, and let
$\nu_{i,\chi}^{a,r}\in\{0,1\}$ equal one when a forbidden effect occurs.
The run-level automation-point completion score and its aggregate are

\begin{equation}
\begin{gathered}
C_{i,\chi}^{a,r}
=
\left(1-\nu_{i,\chi}^{a,r}\right)
\sum_{j\in\mathcal{V}_{i,\chi}}
\beta_{ij,\chi}z_{ij,\chi}^{a,r},
\\[0.8ex]
\operatorname{APComp}_{\chi}(a)
=
\frac{1}{|\mathcal{X}_{\chi}|}
\sum_{i\in\mathcal{X}_{\chi}}
\frac{1}{n_i}
\sum_{r=1}^{n_i}
C_{i,\chi}^{a,r}.
\end{gathered}
\label{eq:automation-point-completion}
\end{equation}

Table~\ref{tab} reports
$\operatorname{APComp}{\mathrm{static}}$ and
$\operatorname{APComp}{\mathrm{dynamic}}$. Each automation point is
binary, while the run-level score is fractional. Static completion
evaluates baseline deliverable points, whereas dynamic completion
evaluates event-transition and post-event adaptation points. These
condition-specific point sets are complementary rather than nested, so
neither aggregate is constrained to be smaller than the other.

\begin{equation}
\operatorname{Diag}_{k}(a)
=
\frac{1}{|\mathcal{X}_{k}|}
\sum_{i\in\mathcal{X}_{k}}
\frac{1}{n_i}
\sum_{r=1}^{n_i}
\frac{1}{|\mathcal{C}_{ik}|}
\sum_{c\in\mathcal{C}_{ik}}p_{ick}^{a,r},
\label{eq:diagnostic}
\end{equation}
where $\mathcal{X}{k}$ contains only episodes to which dimension $k$
applies. Artifact, Evidence, Trace Integrity, Financial Grounding,
Sheet Contract, and Cross-Artifact Consistency use their corresponding
verifier checks. Replan Recovery and Fallback Recovery are restricted
to runs in which replanning or an allowed fallback is required.
Safety Violation is the mean forbidden-effect indicator
$\nu{i,\chi}^{a,r}$.

\section{FinCUABuildAgent}

FinCUABuildAgent operationalizes dynamic financial benchmark construction
as a capability-guided and validation-gated multi-agent pipeline.
Given a preregistered target slot, it first refines the request into
a grounded WorkflowCore and then compiles it into an executable
episode. Figure~\ref{fig:FinCUABuildAgent-overview} summarizes the
construction stages and their realization through specialist agents
and stage-specific tools.

\subsection{FinCUABuildAgent Architecture}
\label{sec:FinCUABuildAgent-architecture}

The Factory Controller coordinates eleven specialist agents through a
generate--audit--repair workflow and routes failed candidates to the
responsible stage. The Benchmark Scout identifies financially meaningful
workflow candidates, which are then formalized by the Workflow Miner and
Process Analyst in terms of objectives, inputs, applications, and
deliverables. The Provenance and Grounding Agent verifies the entity,
reporting period, unit, and financial sources. The Taxonomy and Allocation
Agent assigns each grounded workflow to the benchmark taxonomy, balances
portfolio coverage, and removes semantic duplicates.

Given a grounded WorkflowCore and its target capability, the Capability
and Dependency Compiler specifies the goal conditions, evidence
dependencies, intermediate states, and semantic operations required to
instantiate the capability. The Workflow and Episode Designer integrates
these elements into a user-facing task with artifact, evidence, safety,
and runtime-event specifications. The required operations are then mapped
by the Tool Architect and Synthesizer to an episode-level ToolGraph with
valid fallback paths. The Environment and Dynamics Composer materializes
the financial workspace, initial application state, reset configuration,
and replayable runtime event to produce an executable episode.

The Independent Oracle and Verifier Agent derives the hidden goal state
and verification conditions from grounded evidence before observing any
candidate solution. The Reference Solver executes the task through the
same interface available to evaluated CUA agents, confirming that the
episode remains solvable after the runtime change. The Red-Team and
Calibration Agent tests incorrect financial values, stale evidence,
ignored events, unauthorized effects, and shortcut solutions. The Factory
Controller releases only episodes that pass these checks and returns failed
episodes for targeted revision. Separating ToolGraph specification from
environment materialization and committing the oracle before reference
execution reduce cross-stage leakage and preserve independence between
episode construction and qualification.

\subsection{Typed Agent--Tool Orchestration}
\label{sec:typed-agent-tool-orchestration}

FinCUABuildAgent connects its eleven specialist agents through a typed
orchestration graph
$\mathcal{G}^{\mathrm{O}}=(\mathcal{A},\mathcal{E}^{\mathrm{O}})$, where
$\mathcal{A}={A_{01},\ldots,A_{11}}$. Each agent $A_k$ receives the
validated artifacts $\mathbf{h}_{\operatorname{pa}(k)}$ produced by its
parent agents, invokes a role-specific allowlist of tools
$\mathcal{U}_k$, and returns an artifact $h_k$ together with its
tool-call trace $\ell_k$. The handoff relation $\mathcal{R}_k$ checks
the required artifact type, source lineage, episode version, and
cross-stage dependencies. We activate the downstream stage according to

\begin{equation}
\alpha_k =
\mathbf{1}\!\left[
\left(\mathbf{h}_{\operatorname{pa}(k)},h_k\right)
\in\mathcal{R}_k
;\land;
\ell_k\in\mathsf{Trace}(\mathcal{U}_k)
\right].
\label{eq}
\end{equation}

Here, $\mathsf{Trace}(\mathcal{U}_k)$ contains traces in which every
tool invocation is authorized and satisfies its declared input--output contract. Thus, $\alpha_k=1$ only when both the artifact and its tool-use trace pass validation; otherwise, the failed contract is
returned to the responsible agent for repair.

Agents $A_{01}$--$A_{04}$ handle workflow discovery and grounding.
They identify candidate workflows, retrieve relevant source entities,
and apply licensing and deduplication constraints so that only
consistent and non-redundant WorkflowCores enter the graph.

Agents $A_{05}$--$A_{08}$ compile and materialize executable episodes. They assemble goal predicates, state representations, and capability dependencies, select compatible schemas and runtime-event families, and synthesize interface adapters when needed. Contract, permission, and runtime-conformance checks precede the instantiation of a reproducible environment with a fixed initial state.

Agents $A_{09}$--$A_{11}$ perform qualification and assurance through
separate oracle, reference-execution, and red-team pathways. The oracle inspects hidden goal states and simulates event effects under controlled resets, while the reference solver uses only episode-visible interfaces to produce executable traces. The red-team evaluator probes stale evidence, leakage paths, invalid shortcuts, and unauthorized side effects. Contract or handoff violations trigger rollback and recomputation of the affected downstream artifacts.

\section{Experiments}

\begin{table*}[!t]
\centering
\renewcommand{\arraystretch}{0.6}
\setlength{\tabcolsep}{2.5pt}
\resizebox{\textwidth}{!}{%
\begin{tabular}{lllcccccccccc}
\toprule
Family
& Generation method
& Backbone
& H-Acc $\uparrow$
& L-Acc $\uparrow$
& Trace $\uparrow$
& Exec $\uparrow$
& V-Acc $\uparrow$
& Dyn $\uparrow$
& CapMatch $\uparrow$
& CapCov $\uparrow$
& Dup $\downarrow$
& Qual $\uparrow$ \\
\midrule

Rule-based
& Deterministic templates
& n/a
& 0.0\% & 11.1\% & 74.7\% & 68.4\% & 69.2\%
& 29.6\% & 0.0\% & 0.0\% & 33.3\% & 0.0\% \\
\midrule

\multirow{5}{*}{\shortstack[l]{Direct LLM\\no tools}}
& \multirow{5}{*}{Direct prompting}
& MiniMax-M3
& 1.6\% & 24.3\% & 42.4\% & 46.0\% & 54.2\%
& 36.5\% & 23.6\% & 5.2\% & 28.1\% & 11.3\% \\

&
& GPT-5.5
& 3.2\% & 27.3\% & 38.4\% & 56.6\% & 37.6\%
& 17.4\% & 31.9\% & 3.1\% & 34.6\% & 8.7\% \\

&
& Qwen3.5-397B-A17B
& 2.1\%& 26.5\% & 42.7\% & 57.6\% & 38.9\%
& 43.7\% & 45.4\% & 8.3\% & 45.8\% & 6.8\% \\

&
& Kimi K3
& 6.0\%& 32.3\%& 53.1\% & 52.5\%& 45.2\%
& 20.4\%& 46.9\%& 4.2\%& 45.2\% & 12.2\% \\

&
& MiMo-V2.5-Pro
& 2.4\% & 25.0\% & 39.6\% & 55.6\% & 37.5\%
& 52.5\% & 35.7\% & 5.0\% & 41.8\% & 12.3\% \\
\midrule

\multirow{4}{*}{\shortstack[l]{Direct LLM\\+ finance tools}}
& \multirow{4}{*}{\shortstack[c]{Tool-augmented\\prompting}}

& MiniMax-M3
& 12.9\% & 21.1\% & 67.1\% & 57.5\% & 54.2\%
& 38.8\% & 56.9\% & 31.6\% & 37.4\% & 17.5\% \\

&
& Qwen3.5-397B-A17B
& 13.5\% & 25.1\% & 65.2\% & 52.3\% & 38.9\%
& 43.7\% & 51.3\% & 35.8\% & 40.8\% & 12.8\% \\

&
& Kimi K3
& 14.1\% & 25.0\% & 62.1\% & 56.2\%& 43.6\%
& 37.3\%& 49.7\%& 30.0\%& 38.1\%& 10.9\% \\

&
& MiMo-V2.5-Pro
& 12.1\% & 35.9\% & 63.1\% & 57.6\% & 38.9\%
& 43.6\% & 45.8\% & 28.3\% & 46.5\% & 10.8\% \\
\midrule

\multirow{4}{*}{\shortstack[l]{Agentic\\baselines}}
& Adapted AutoBencher
& MiniMax-M3
& 12.0\% & 10.4\% & 11.1\% & 11.1\% & 10.9\%
& 10.9\% & 10.4\% & 37.5\% & 64.1\% & 8.3\% \\

& Adapted BENCHAGENTS
& MiniMax-M3
& 5.2\% & 5.0\% & 5.0\% & 5.0\% & 5.0\%
& 5.0\% & 5.0\% & 12.5\% & 65.5\% & 1.3\% \\

& Adapted Benchmark Agent
& MiniMax-M3
& 7.8\% & 7.6\% & 8.0\% & 8.0\% & 7.8\%
& 8.0\% & 7.5\% & 25.0\% & 63.0\% & 4.5\% \\

& Single Agent + finance tools
& MiniMax-M3
& 6.8\% & 7.3\% & 7.3\% & 7.3\% & 7.3\%
& 7.3\% & 7.3\% & 25.0\% & 64.3\% & 6.9\% \\
\midrule

\multirow{4}{*}{\textbf{Ours}}
& \multirow{4}{*}{\textbf{FinCUABuildAgent Multi-Agent}}
& MiniMax-M3
& 33.4\% & 55.7\% & 89.1\%
& 73.8\% & 72.3\% & 70.0\%
& 78.5\% & 50.7\% & 12.5\%
& 31.3\% \\

&
& Qwen3.5-397B-A17B
& 30.4\% & 53.1\% & 90.3\%
& 73.1\% & 64.7\% & 90.3\%
& 67.9\% & 58.3\% & 12.2\%
& 41.9\% \\

&
& Kimi K3
& 35.2\%& 50.8\%& 85.4\%
& 65.9\%& 71.7\%& 85.5\%
& 66.1\%& 58.3\%& 8.9\%
& 40.3\% \\

&
& MiMo-V2.5-Pro
& 28.6\% & 49.7\% & 89.3\%
& 81.8\% & 78.1\% & 72.6\%
& 70.5\% & 44.8\% & 11.9\%
& 34.9\% \\
\bottomrule
\end{tabular}%
}

\caption{
Benchmark-generation quality across financial CUA construction methods.
Agentic baselines use MiniMax-M3 under matched construction settings.
Values are averaged over the preregistered construction runs and then
macro-averaged across WorkflowCores.
}
\label{tab:benchgen-quality}
\end{table*}

\subsection{Experimental Setup}

\textbf{Benchmarks.} 
We evaluate benchmark construction on FinCUABuildBench, which contains 576 target slots across 24 financial workflow categories and three dynamic profiles: State Update, Tool Failure, and Policy Change. To assess the downstream diagnostic value of generated tasks, we evaluate CUA agents on a balanced subset of 144 qualified episodes, with three independent runs per agent. We further test construction transfer on 90 held-out requests covering three unseen workflow scenarios: FP\&A Variance Investigation, AP Exception Handling, and Portfolio Risk \& Compliance.

\textbf{Baseline methods.} We compare FinCUABuildAgent against rule-based templates, LLM prompting (with and without tools), tool-augmented prompting, and adapted benchmark-generation systems including AutoBencher \cite{li2025autobencher}, BENCHAGENTS \cite{butt2024benchagents}, and Benchmark Agent \cite{xiong2026benchmarkagent}, as well as a single-agent tool-using baseline. We evaluate multiple backbones including MiniMax-M3, GPT-5.5, Qwen3.5-397B-A17B, Kimi-K3, and MiMo-V2.5-Pro depending on the setting. 

Downstream evaluation on the 144-task subset covers general-purpose agents, financial agents, and controlled baselines. 
We also report held-out transfer and ablations against a budget-matched collapsed-controller variant and component removals. The collapsed controller in Table~\ref{tab:ablation-analysis} is an internal topology ablation and is distinct from the external single-agent baseline in Table~\ref{tab:benchgen-quality}.

\textbf{Evaluation protocol.} All methods share the same target specifications, resource budgets, and evaluation interfaces. Tool access follows the declared baseline setting and is held fixed within each matched comparison; no method has access to gold ToolGraphs or reference trajectories. The primary metric is Qualified Task Rate (Qual), the fraction of target slots that pass the strict conjunction of every preregistered episode-level hard gate and mandatory constraint in $\Gamma$. H-Acc is a stratified human audit and is not included in this conjunction; V-Acc is reported as a mean verifier score, while strict qualification uses its thresholded episode-level form. We further report L-Acc, Trace, Exec, V-Acc, Dyn, CapMatch, CapCov, and Dup as diagnostic metrics. Construction metrics are macro-averaged over WorkflowCores. Human and LLM-based assessments are method-blind, while execution-derived metrics are computed automatically from frozen artifacts and traces. Full metric definitions are provided in the supplementary material.

\subsection{Main Result}

Table~\ref{tab:benchgen-quality} shows that FinCUABuildAgent provides the
strongest overall profile on the coupled construction metrics. Finance
tools alone consistently improve provenance and capability coverage,
but their effects on dynamicity, verifier reliability, and strict
qualification remain uneven. Across the four backbones shared with
tool-augmented prompting, FinCUABuildAgent raises Dyn by 29.0--48.2
percentage points, reduces Dup by 24.9--34.6 points, and improves
strict Qual by 13.8--29.4 points. Qwen3.5 achieves the highest Qual at
41.9\%. The adapted agentic baselines reach only 1.3--8.3\% Qual under
MiniMax-M3, indicating that generic benchmark-generation pipelines do
not directly recover the coupled financial grounding, executable
state, runtime event, and verification contracts required here.

\subsection{Diagnostic Effectiveness of Generated Tasks}

\begin{table*}[!t]
\centering
\renewcommand{\arraystretch}{1.12}
\resizebox{\textwidth}{!}{
\begin{tabular}{llccc ccc ccccc}
\toprule
\multirow{2}{*}{Category}
&
\multirow{2}{*}{Evaluated Agent / Configuration}
&
\multicolumn{3}{c}{Output and Evidence Quality}
&
\multicolumn{3}{c}{Task Performance}
&
\multicolumn{5}{c}{Dynamic Capability Diagnosis}
\\
\cmidrule(lr){3-5}
\cmidrule(lr){6-8}
\cmidrule(lr){9-13}

&
&
\makecell{Artifact $\uparrow$}
&
\makecell{Evidence $\uparrow$}
&
\makecell{Trace\\Integrity $\uparrow$}
&
\makecell{Static\\Completion $\uparrow$}
&
\makecell{Dynamic\\Completion $\uparrow$}
&
\makecell{Financial\\Grounding $\uparrow$}
&
\makecell{Sheet\\Contract $\uparrow$}
&
\makecell{Cross-Artifact\\Consistency $\uparrow$}
&
\makecell{Replan\\Recovery $\uparrow$}
&
\makecell{Fallback\\Recovery $\uparrow$}
&
\makecell{Safety\\Violation $\downarrow$}
\\
\midrule

\multirow{5}{*}{\makecell[l]{Direct MiniMax\\protocols}}
& Text-only
& 0.709 & 0.000 & 0.384
& 0.320 & 0.081 & 0.000 & 0.610& 0.018 & 0.132 & 0.147 & 0.153 \\

& Artifact draft
& 0.687 & 0.000 & 0.384
& 0.326 & 0.283 & 0.167 & 0.610 & 0.084 & 0.631 & 0.430 & 0.100 \\

& Structured evidence
& 0.719 & 0.493 & 0.257
& 0.520 & 0.069& 0.929 & 0.608 & 0.217 & 0.151 & 0.100& 0.009 \\

& Structured + verifier repair
& 0.721 & 0.631 & 0.320
& 0.532 & 0.000& 0.930 & 0.611 & 0.280 & 0.000 & 0.000 & 0.023 \\

& External structured agent
& 0.722 & 0.620 & 0.149
& 0.524 & 0.676& 0.941 & 0.628 & 0.130 & 0.771 & 0.771 & 0.021 \\

\midrule

\multirow{5}{*}{\makecell[l]{Native general-agent\\frameworks}}
& AutoGen GroupChat
& 0.701 & 0.159 & 0.033
& 0.641 & 0.043& 1.000 & 0.739 & 0.154 & 0.277 & 0.230 & 0.035 \\

& Magentic-One
& 0.702 & 0.219 & 0.045
& 0.646 & 0.004& 0.998 & 0.749 & 0.211 & 0.027 & 0.022 & 0.035 \\

& CrewAI
& 0.709 & 0.101 & 0.021
& 0.660 & 0.026& 0.994 & 0.812 & 0.090 & 0.169 & 0.141 & 0.021 \\

& LangGraph
& 0.690 & 0.096 & 0.022
& 0.622 & 0.041& 0.993 & 0.731 & 0.089 & 0.262 & 0.219 & 0.035 \\

& CAMEL
& 0.684 & 0.116 & 0.025
& 0.623 & 0.033& 0.997 & 0.711 & 0.110 & 0.212 & 0.174 & 0.028 \\

\midrule

\multirow{5}{*}{\makecell[l]{Financial-agent\\candidates}}
& FinRobot
& 0.684 & 0.341 & 0.005
& 0.667 & 0.000& 1.000 & 0.695 & 0.005 & 0.000 & 0.000 & 0.319 \\

& TradingAgents
& 0.692 & 0.366 & 0.006
& 0.639 & 0.000& 1.000 & 0.588 & 0.005 & 0.000 & 0.000 & 0.241 \\

& FinTeam
& 0.689 & 0.397 & 0.009
& 0.479 & 0.224& 0.982 & 0.619 & 0.006 & 0.417 & 0.132 & 0.403 \\

& FinGPT
& 0.689 & 0.400 & 0.004
& 0.583 & 0.364& 1.000 & 0.703 & 0.003 & 0.250 & 0.104 & 0.329 \\

& FinRL
& 0.690 & 0.387 & 0.008
& 0.643 & 0.015& 1.000 & 0.674 & 0.007 & 0.025 & 0.015& 0.227 \\

\midrule

\multirow{2}{*}{\makecell[l]{Matched agent\\controls}}
& Genuine SingleController
& 0.998 & 0.995 & 0.451
& 0.997 & 0.258 & 0.995 & 0.998 & 0.995 & 0.613 & 0.437 & 0.000 \\

& ReAct
& 0.606 & 0.035 & 0.013
& 0.375 & 0.228& 0.496 & 0.571 & 0.004 & 0.556 & 0.411 & 0.005 \\

\bottomrule
\end{tabular}
}
\caption{
Diagnostic evaluation on 144 qualified tasks sampled from FinCUABuildBench. Each agent or configuration runs all 144 tasks with three independent replicates, yielding 432 trajectories. Static and Dynamic Completion are mean condition-specific automation-point completion scores computed from the same trajectories over complementary, non-nested point sets; the remaining columns average their applicable verifier checks.
}
\label{tab}
\end{table*}

Table~\ref{tab} evaluates whether the qualified tasks constructed by FinCUABuildAgent provide meaningful diagnostic signals for downstream CUAs. Since the frozen reference solver succeeds on every selected task while evaluated agents obtain non-saturated and differentiated scores, the generated tasks are both feasible and discriminative. Native general-agent frameworks achieve 0.622--0.660 Static Completion and near-perfect financial grounding but only 0.004--0.043 Dynamic Completion, showing that grounding and general tool use do not ensure adaptation to runtime events. The external structured configuration reaches the highest Dynamic Completion (0.676), confirming that stronger performance is attainable; its Dynamic score exceeds its Static score because the metrics use different, non-nested automation-point sets. Meanwhile, the matched SingleController combines near-perfect artifact, evidence, static-completion, and cross-artifact scores with only 0.258 Dynamic Completion. Together, these results show that the generated tasks expose complementary failure modes in financial grounding, task completion, and runtime adaptation.

\subsection{Generalization to Held-Out Workflows}

Table~\ref{tab:heldout-generalization} shows uneven transfer to held-out
WorkflowCores within the covered workflow taxonomy. V-Acc remains at
least 86.3\% in every workflow--profile cell, whereas only 13.3\% of
episodes satisfy the strict qualification criterion. The gap is most
pronounced under P3: 46.7\% of the episodes are executable, but only
6.7\% strictly qualify. This suggests that the independent oracle
transfers reliably from frozen evidence and typed goal predicates,
while coupling a policy change to a replayable environment and a
capability-essential post-event solution remains difficult. FP\&A
achieves higher aggregate Exec (56.7\%) and Qual (20.0\%) than AP
exception handling (43.3\%, 10.0\%) and portfolio risk and compliance
(33.3\%, 10.0\%), although the latter attains higher CapMatch. Overall,
the results support partial within-taxonomy WorkflowCore transfer,
without implying generalization to entirely unseen workflow categories.

\begin{table}[t]
\centering
\setlength{\tabcolsep}{4pt}
\renewcommand{\arraystretch}{1.15}

\resizebox{\columnwidth}{!}{
\begin{tabular}{l l ccccc}
\toprule
\textbf{Scenario} & \textbf{Profile} & \textbf{Exec $\uparrow$} & \textbf{V-Acc $\uparrow$} & \textbf{Dyn $\uparrow$} & \textbf{CapMatch $\uparrow$} & \textbf{Qual $\uparrow$} \\
\midrule

\multirow{3}{*}{\shortstack[c]{FP\&A Variance\\Investigation}}
& P1: State Update   & 60.0\% & 95.8\%  & 40.0\% & 40.0\% & 30.0\% \\
& P2: Tool Failure    & 50.0\% & 90.0\%  & 30.0\% & 20.0\% & 20.0\% \\
& P3: Policy Change   & 60.0\% & 100.0\% & 30.0\% & 10.0\% & 10.0\% \\

\midrule

\multirow{3}{*}{\shortstack[c]{AP Exception\\Handling}}
& P1: State Update   & 40.0\% & 95.8\%  & 20.0\% & 20.0\% & 10.0\% \\
& P2: Tool Failure    & 50.0\% & 100.0\% & 50.0\% & 30.0\% & 20.0\% \\
& P3: Policy Change   & 40.0\% & 88.7\%  & 20.0\% & 30.0\% & 0.0\%  \\

\midrule

\multirow{3}{*}{\shortstack[c]{Portfolio Risk\\\& Compliance}}
& P1: State Update   & 30.0\% & 95.8\%  & 20.0\% & 50.0\% & 20.0\% \\
& P2: Tool Failure    & 30.0\% & 86.3\%  & 30.0\% & 20.0\% & 0.0\%  \\
& P3: Policy Change   & 40.0\% & 96.1\%  & 30.0\% & 40.0\% & 10.0\% \\

\bottomrule
\end{tabular}
}

\caption{
Held-out workflow generalization under a frozen FinCUABuildAgent
configuration. Each of three unseen financial scenarios contains ten
held-out WorkflowCores instantiated once under P1--P3, yielding 90
episodes.}
\label{tab:heldout-generalization}
\end{table}

\subsection{Component and Topology Ablations}

\begin{table}[t]
\centering
\setlength{\tabcolsep}{3pt}
\renewcommand{\arraystretch}{1.15}

\resizebox{\columnwidth}{!}{
\begin{tabular}{l ccccccc}
\toprule
\textbf{Variant}
& \textbf{L-Acc $\uparrow$}
& \textbf{Trace $\uparrow$}
& \textbf{Exec $\uparrow$}
& \textbf{V-Acc $\uparrow$}
& \textbf{Dyn $\uparrow$}
& \textbf{CapCov $\uparrow$}
& \textbf{Dup $\downarrow$}
\\
\midrule

\textbf{FinCUABuildAgent Multi-Agent}
& 55.7\% & 89.1\% & 73.8\% & 72.3\% & 70.0\% & 50.7\% & 12.5\% \\

Collapsed controller (budget matched)
& 26.0\% & 22.9\% & 18.8\% & 40.1\%  & 12.5\% & 27.3\% & 49.7\% \\

w/o Grounding \& Provenance
& 37.5\% & 79.0\% & 10.4\% & 88.5\% & 38.4\% & 0.0\%  & 25.0\% \\

w/o Dependency Compiler
& 0.0\%  & 40.5\% & 8.6\%  & 0.0\%  & 0.0\%  & 0.0\%  & 100.0\% \\

w/o Tool Architect
& 0.0\%  & 3.2\%  & 0.0\%  & 1.7\%  & 0.6\%  & 0.0\%  & 100.0\% \\

w/o Dynamics Composer
& 1.0\%  & 10.2\% & 1.6\%  & 1.0\%  & 5.8\%  & 1.6\%  & 96.7\% \\

w/o Independent Oracle
& 8.9\%  & 61.6\% & 58.4\% & 52.0\% & 63.2\% & 35.9\% & 43.2\% \\

w/o Red Team \& Repair
& 0.5\%  & 78.9\% & 69.8\% & 68.8\% & 73.2\% & 39.6\% & 20.8\% \\

\bottomrule
\end{tabular}
}

\caption{
Component and topology ablations of FinCUABuildAgent with MiniMax-M3 on the
same 576 target slots as the corresponding full-system row in
Table~\ref{tab:benchgen-quality}. The collapsed-controller variant is
budget matched; all rows follow the same construction-run aggregation
and scoring protocol.
}
\label{tab:ablation-analysis}
\end{table}

Table~\ref{tab:ablation-analysis} contrasts diagnostic failure patterns. Under the matched budget, staged
coordination raises Exec from 18.8\% for the collapsed controller to
73.8\%, showing that typed handoffs reduce inconsistencies among goal
predicates, ToolGraphs, environment states, and verification logic. Removing Grounding \& Provenance yields 88.5\% V-Acc but only 10.4\% Exec, showing that verifier accuracy alone cannot ensure a grounded, executable task portfolio.
Removing the Dependency Compiler, Tool Architect, or Dynamics Composer
yields Dup of at least 96.7\% and CapCov of at most 1.6\%, identifying
these stages as the contract spine from semantic intent to executable
dynamics. Removing the Independent Oracle reduces V-Acc from 72.3\% to
52.0\% and L-Acc from 55.7\% to 8.9\%, while raising Dup from 12.5\% to
43.2\%. Removing red-team repair preserves several structural metrics
but reduces L-Acc to 0.5\% and raises Dup to 20.8\%. These patterns show
complementary stage contributions.

\section{Conclusion}

We introduced FinCUABuildBench, a 576-slot benchmark for constructing dynamic financial CUA tasks across 24 workflow categories and three runtime-dynamic profiles, together with FinCUABuildAgent, a capability-guided multi-agent framework for producing grounded, executable, replayable, and verifiable episodes. Against matched tool-augmented prompting across four backbones, FinCUABuildAgent improves Dyn by 29.0--48.2 points and strict Qual by 13.8--29.4 points while reducing Dup by 24.9--34.6 points; Qwen3.5 achieves the highest Qual of 41.9\%. 
The generated qualified tasks yield non-saturated and differentiated automation-point completion scores across existing agents, demonstrating their downstream diagnostic value.
Dynamic Completion is lower than Static Completion in 16 of 17 configurations, revealing limited adaptation to post-start changes. Held-out evaluation reaches 13.3\% strict Qual and shows partial transfer within the covered taxonomy, while ablations support staged multi-agent coordination. Together, these results frame benchmark construction as a systems problem requiring joint control over grounding, state, dynamics, and verification. Overall, validation-gated construction provides a practical basis for dynamic financial CUA evaluation, with remaining challenges in qualification yield, broader transfer, production coverage, source licensing, and verifier maintenance.

\bibliography{references.bib}



\section{Benchmark-Generation Metric Definitions}

Let $m\in\mathcal{M}$ denote a generation method, let
$N_m^{\star}$ be its preregistered number of target episode slots, and
let $\mathcal{B}_m=\{x_i\}_{i=1}^{N_m^{\star}}$ contain the candidate
assigned to each slot in one construction run. An unfilled, malformed,
or non-materialized slot is represented by an invalid candidate and
receives zero on downstream hard gates. The definitions below specify
one run; when the preregistered protocol contains multiple independent
construction runs, the reported value is the mean of the run-level
metric. Unless stated otherwise, metrics lie in $[0,1]$, are
macro-averaged across WorkflowCores, and are converted to percentages
in Table~\ref{tab:benchgen-quality}.

\paragraph{Human Acceptance Rate (H-Acc).}
Financial-domain reviewers assess financial correctness, workflow realism, cross-component consistency, and alignment with the target capability. Let $h_{ir}\in\{0,1\}$ be reviewer $r$'s decision for episode $i$, and let $\mathcal{H}_m$ be a stratified, method-blind sample of target slots, with invalid slots treated as rejected. We define
\begin{equation}
\operatorname{HAcc}(m)=
\frac{1}{|\mathcal{H}_m|}
\sum_{i\in\mathcal{H}_m}
\mathbf{1}\!\left[
\frac{1}{|\mathcal{R}_i|}
\sum_{r\in\mathcal{R}_i}h_{ir}\geq\tau_{\mathrm H}
\right].
\label{eq:hacc}
\end{equation}
Because H-Acc is estimated on a stratified sample rather than on every
target slot, it is reported as a method-blind external audit and is not
included in the strict episode-level qualification indicator.

\paragraph{LLM-as-Judge Acceptance Rate (L-Acc).}
A method-blind judge receives the workflow seed, target capability, generated instruction, source/state summary, WorkflowGraph, ToolGraph, runtime event, and goal/verifier synopsis. For judge $j$ and rubric dimension $d$, let $s_{ijd}\in[1,5]$ be the score, $w_d$ its preregistered weight, and $\bar{s}_{id}$ the mean across judges. Let
\[
\ell_i
=
\mathbf{1}\!\left[
\min_{d\in\mathcal D}\bar{s}_{id}\geq\tau_{\min}
\land
\sum_{d\in\mathcal D}w_d\bar{s}_{id}\geq\tau_{\mathrm L}
\right].
\]
We use
\begin{equation}
\operatorname{LAcc}(m)
=
\frac{1}{N_m^\star}
\sum_{i=1}^{N_m^\star}\ell_i.
\label{eq:lacc}
\end{equation}
The minimum-dimension constraint prevents a strong average from hiding a critical failure.

\paragraph{Source \& State Traceability Rate (Trace).}
Let $\mathcal{P}_i$ be the required provenance atoms for episode $i$, including source identifier, version or accession, hash, period, concept, unit, initial-state hash, event-state transition, goal predicate, and verifier reference. With $\operatorname{valid}_i(p)$ denoting schema, referential, and hash validity, define
\[
t_i
=
\prod_{p\in\mathcal{P}_i}
\mathbf{1}[\operatorname{valid}_i(p)=1].
\]
\begin{equation}
\operatorname{Trace}(m)=
\frac{1}{N_m^{\star}}
\sum_{i=1}^{N_m^{\star}}
t_i.
\label{eq:trace}
\end{equation}

\paragraph{Execution \& Replay Success Rate (Exec).}
Let $b_i$ and $g_i$ indicate successful environment boot and ToolGraph execution. Let $r_{i\ell}$, $o_{iq}$, and $\eta_{ik}$ indicate reset run $\ell$, reference-solver run $q$, and event replay $k$. Define
\begin{gather}
\rho_i^{\mathrm{rep}}
=
\left(\prod_{\ell=1}^{L} r_{i\ell}\right)
\left(\prod_{q=1}^{Q} o_{iq}\right)
\left(\prod_{k=1}^{K_i} \eta_{ik}\right),
\label{eq:replay-success}
\\
\displaystyle
e_i
=
b_i g_i \rho_i^{\mathrm{rep}},
\\
\operatorname{Exec}(m)
=
\frac{1}{N_m^\star}
\sum_{i=1}^{N_m^\star}
e_i.
\label{eq:exec}
\end{gather}
Static controls set the event-replay product to one.

\paragraph{Verifier Accuracy (V-Acc).}
Let $\mathcal{Z}_i^{-}$ be semantically incorrect mutations and $\mathcal{Z}_i^{+}$ be valid alternative solutions. For hidden verifier $V_i(z)\in\{0,1\}$, define
\begin{gather}
\operatorname{VMR}_i
=
\frac{1}{|\mathcal{Z}_i^{-}|}
\sum_{z\in\mathcal{Z}_i^{-}}
\mathbf{1}[V_i(z)=0],
\\
\operatorname{VAA}_i
=
\frac{1}{|\mathcal{Z}_i^{+}|}
\sum_{z\in\mathcal{Z}_i^{+}}
\mathbf{1}[V_i(z)=1],
\\
\displaystyle
v_i
=
\mathbf{1}\!\left[
\frac{\operatorname{VMR}_i+\operatorname{VAA}_i}{2}
\geq\tau_{\mathrm V}
\right],
\\
\operatorname{VAcc}(m)
=
\frac{1}{N_m^\star}
\sum_{i=1}^{N_m^\star}
\frac{
\operatorname{VMR}_i+\operatorname{VAA}_i
}{2}.
\label{eq:vacc}
\end{gather}

A missing verifier or missing required test set contributes zero and
sets $v_i=0$. Thus, the reported V-Acc column is a continuous mean,
while strict qualification uses the thresholded indicator $v_i$.

\paragraph{Valid Dynamic Task Rate (Dyn).}
Let $\mathcal{D}_m$ be the planned dynamic slots. For $i\in\mathcal{D}_m$, $t_i^{\mathrm{evt}}>t_i^0$ indicates a post-start event; $r_i^{\mathrm{evt}}$ indicates deterministic replay; $(\pi_i^{-},G_i^{-})$ and $(\pi_i^{+},G_i^{+})$ are the valid plan and goal before and after the event; and $y_i^{\mathrm{ignore}}$ and $y_i^{\mathrm{adapt}}$ are event-ignoring and event-adaptive outcomes. We define
\begin{gather}
c_i^{\mathrm{evt}}
=
\mathbf{1}\!\left[
t_i^{\mathrm{evt}}>t_i^0
\land
(\pi_i^{-},G_i^{-})\neq(\pi_i^{+},G_i^{+})
\right],
\label{eq:event-change}
\\
\scalebox{0.88}{$\displaystyle
d_i^{\mathrm{dyn}}
=
c_i^{\mathrm{evt}} r_i^{\mathrm{evt}}
\mathbf{1}\!\left[
V_i(y_i^{\mathrm{ignore}})=0
\land
V_i(y_i^{\mathrm{adapt}})=1
\right]
$},
\label{eq:valid-dynamic}
\\
\operatorname{Dyn}(m)
=
\frac{1}{|\mathcal{D}_m|}
\sum_{i\in\mathcal{D}_m}d_i^{\mathrm{dyn}}.
\label{eq:dyn}
\end{gather}

\paragraph{Target Capability Match Rate (CapMatch).}
Let $a_{ic}\in\{0,1\}$ indicate that episode $i$ targets capability $c$, and let $j_{ic}$ be the method-blind judge's alignment decision. Define the capability-removal gap $\Delta_{ic}=\widehat p_i^{\mathrm{full}}-\widehat p_i^{-c}$ and shortcut success rate $s_i=\widehat p_i^{\mathrm{shortcut}}$. For preregistered margins $\delta_c$ and $\epsilon$,
\[
c_i
=
\prod_{c:a_{ic}=1}
j_{ic}\,
\mathbf{1}\!\left[
\Delta_{ic}\geq\delta_c
\land
s_i\leq\epsilon
\right],
\]
where each target slot has at least one preregistered capability and an
empty or malformed assignment sets $c_i=0$. We define
\begin{equation}
\operatorname{CapMatch}(m)=
\frac{1}{N_m^\star}
\sum_{i=1}^{N_m^\star}c_i.
\label{eq:capmatch}
\end{equation}

\paragraph{Duplicate Task Rate (Dup).}
For two episodes, we combine similarities over instruction text, WorkflowGraph, ToolGraph, goal/verifier structure, and source identity. Let $\mathcal{K}=\{\mathrm{text},\mathrm{wf},\mathrm{tool},\mathrm{goal},\mathrm{source}\}$ and let $\lambda_k$ be nonnegative weights summing to one:
\begin{gather}
S_{ij}
=
\sum_{k\in\mathcal{K}}
\lambda_k S_{ij}^{(k)},
\qquad
\sum_{k\in\mathcal{K}}\lambda_k=1,
\label{eq:similarity}
\\
z_i
=
\mathbf{1}\!\left[
\max_{j\in\mathcal{N}_m(i)\setminus\{i\}}
S_{ij}
\geq
\tau_{\mathrm{dup}}
\right],
\label{eq:duplicate-indicator}
\\
\operatorname{Dup}(m)
=
\frac{1}{N_m^\star}
\sum_{i=1}^{N_m^\star}z_i.
\label{eq:dup}
\end{gather}
Here, $\mathcal{N}_m(i)$ denotes the preregistered comparison
set for episode $i$ among the $N_m^\star$ valid outputs generated
by method $m$, excluding the episode itself. An episode is marked
as duplicate if it has at least one comparison whose composite
similarity reaches $\tau_{\mathrm{dup}}$. Comparisons use
structure-aware dimensions rather than surface text alone.
The comparison order is fixed before evaluation; candidates within a WorkflowCore are compared through structure-aware dimensions rather than text alone.

\paragraph{Strict Qualification Indicator.}
Let
\[
\resizebox{0.92\linewidth}{!}{
$
\mathcal{G}_{\mathrm{hard}}
=
\{\mathrm{LAcc},\mathrm{Trace},\mathrm{Exec},
\mathrm{VPass},\mathrm{Dyn},\mathrm{CapMatch},
\mathrm{Unique}\}
$
}
\]
be the preregistered episode-level hard-gate set. The corresponding
binary indicators are $\ell_i$, $t_i$, $e_i$, $v_i$,
$d_i^{\mathrm{dyn}}$, $c_i$, and $1-z_i$, as defined above. Let
$\gamma_i\in\{0,1\}$ indicate satisfaction of every mandatory safety,
privacy, permission, resource, materialization, and forbidden-effect
constraint in $\Gamma$. The strict episode-level qualification
indicator is
\begin{equation}
q_i
=
\gamma_i\,
\ell_i t_i e_i v_i
d_i^{\mathrm{dyn}}c_i(1-z_i).
\label{eq:qualification-indicator}
\end{equation}
A missing, malformed, or non-materialized component sets its
corresponding gate to zero.

\paragraph{Capability Coverage Rate (CapCov).}
Let $\mathcal{C}$ be the preregistered capability ontology, $\rho_c$ the
required number of strictly qualified episodes for capability $c$,
$\pi_c>0$ its importance weight, and $q_i$ the strict qualification
indicator defined in Eq.~\eqref{eq:qualification-indicator}. Weighted
coverage is
\begin{equation}
\operatorname{CapCov}(m)=
\frac{
\sum_{c\in\mathcal{C}}\pi_c
\mathbf{1}[\sum_iq_i a_{ic}\geq\rho_c]
}{
\sum_{c\in\mathcal{C}}\pi_c
}.
\label{eq:capcov}
\end{equation}

\paragraph{Qualified Task Rate (Qual).}
Using the strict episode-level qualification indicator defined in
Eq.~\eqref{eq:qualification-indicator}, we define

\begin{equation}
\operatorname{Qual}(m)
=
\frac{1}{N_m^\star}
\sum_{i=1}^{N_m^\star} q_i.
\label{eq:qual}
\end{equation}

Qual is the primary end-to-end construction metric, while the
remaining metrics diagnose individual aspects of benchmark quality.
\end{document}